\documentclass[letterpaper, 10 pt, conference]{ieeeconf}  

\IEEEoverridecommandlockouts                              

\usepackage{graphics} 
\usepackage{epsfig} 
\usepackage{times} 
\usepackage{amsmath} 
\usepackage{amssymb}  
\usepackage[tight,footnotesize]{subfigure}
\usepackage{enumerate}
\usepackage{cite}
\usepackage{algorithm}
\usepackage{algpseudocode}
\usepackage{amsmath}
\usepackage{graphics}
\usepackage{epsfig}
\usepackage{amsmath,amssymb}
\usepackage{mathrsfs}
\usepackage[colorlinks,linkcolor=red]{hyperref}

\title{\LARGE \bf
End-to-End 3D Point Cloud Learning for Registration Task Using Virtual Correspondences
}

\author{Zhijian~Qiao, Huanshu~Wei, Zhe~Liu, Chuanzhe~Suo, Hesheng~Wang
}

\begin{document}

\maketitle
\thispagestyle{empty}
\pagestyle{empty}

\begin{abstract}
3D Point cloud registration is still a very challenging topic due to the difficulty in finding the rigid transformation between two point clouds with partial correspondences, and it's even harder in the absence of any initial estimation information. In this paper, we present an end-to-end deep-learning based approach to resolve the point cloud registration problem. Firstly, the revised LPD-Net is introduced to extract features and aggregate them with the graph network. Secondly, the self-attention mechanism is utilized to enhance the structure information in the point cloud and the cross-attention mechanism is designed to enhance the corresponding information between the two input point clouds. Based on which, the virtual corresponding points can be generated by a soft pointer based method, and finally, the point cloud registration problem can be solved by implementing the SVD method. Comparison results in ModelNet40 dataset validate that the proposed approach reaches the state-of-the-art in point cloud registration tasks and experiment resutls in KITTI dataset validate the effectiveness of the proposed approach in real applications.Our source code is available at \url{https://github.com/qiaozhijian/VCR-Net.git}
\end{abstract}

\section{Introduction}

\begin{figure}
	\centering
	\includegraphics[width=\columnwidth]{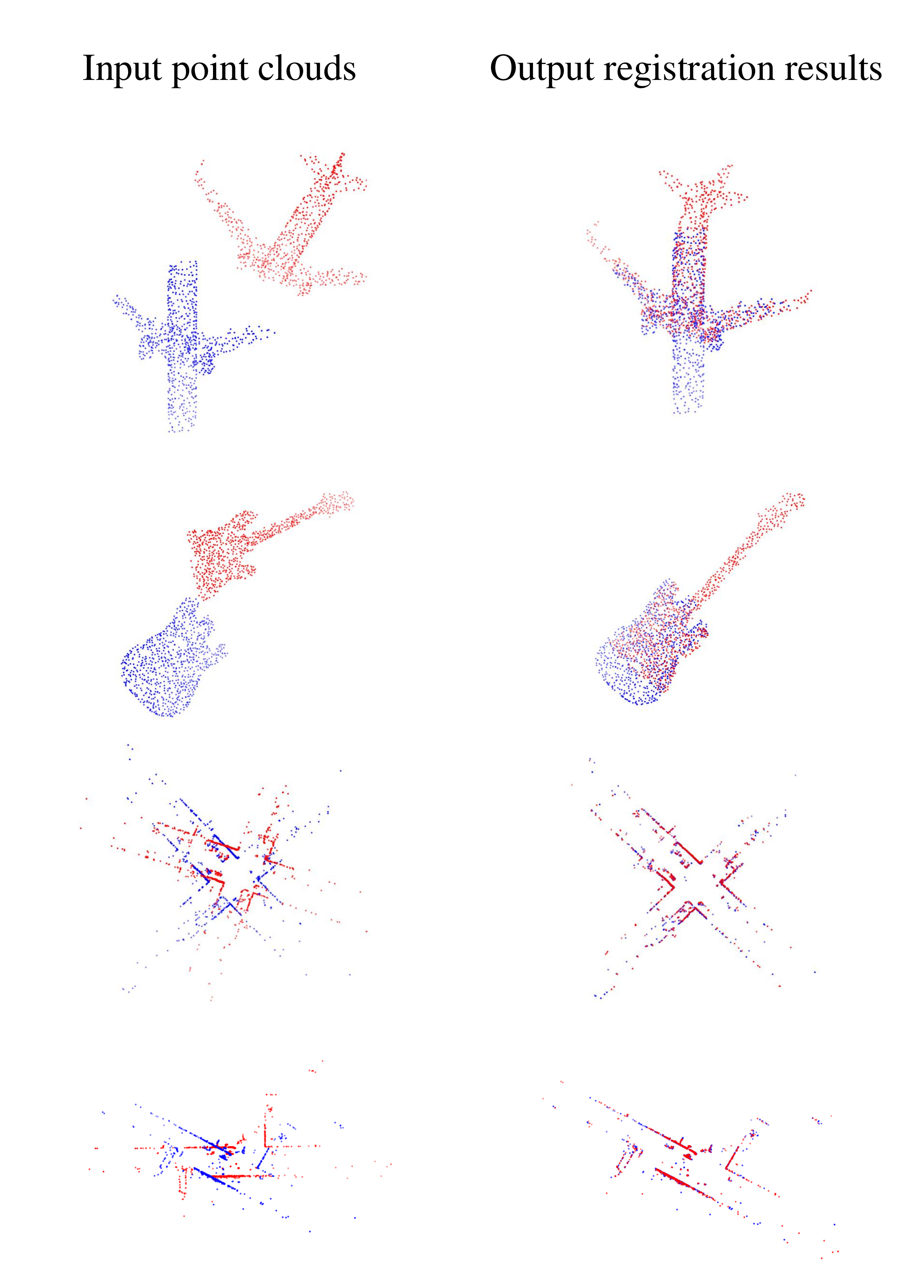}
	\caption{Point cloud registration with partial correspondences: Finding the rigid transformation between two input point clouds with partial correspondences.}
	\label{fig:system}
\end{figure}
Point cloud registration is a highly critical task in self-driving and robotic applications as it estimates the relative transformation between the two frames, which is essential for the object detection and tracking as well as the pose estimation and localization tasks. In the past few decades, vision-based methods have been widely studied and can be mainly divided into two categories, model-based \cite{MurORB} and learning-based \cite{LiUnDeepVO}. Unfortunately, these vision-based solutions are not reliable in practical application as it is sensitive to season, illumination and viewpoint variations, and may entirely break down in dark environments. Recently, 3D point cloud-based methods have drawn significant attention due to its robustness to changing environments and extreme light conditions \cite{liu2019seqlpd:, 2020Deep}. The task of aligning several point clouds by estimating the relative transformation between them is known as the point cloud registration or the pose estimation \cite{Paul1992A}. Considering the practical application requirements in self-driving, the point cloud registration task is challenging due to the characteristics of lidar and the application requirements in large-scale scenarios. First, the sparsity of the point cloud makes it infeasible to always find correspondences between two frames. Second, the processing of millions of points requires effective algorithms and powerful computing hardware. Third, the large size of the scene aggravates the sparsity of the point cloud, along with the non-uniform point density, increases the difficulty of finding correspondences.

Traditional point cloud registration methods \cite{Paul1992A,Jiaolong2013Go} start with a coarse initial value which usually comes from odometers and iterate until the optima is reached. Whereas, as the question has been proven to be non-convex \cite{WangDeep}, the result is highly dependent to a good initial value, which means an inappropriate initial value would lead to a local optima. Nowadays, deep learning-based methods have been proved to have the potential to make a significant improvement for most semantic computer vision tasks, such as classification, detection, segmentation and registration \cite{ZhouVoxelNet,SindagiMVX,Garcia2016PointNet}. 
However, how to extract efficient features of the point clouds and how to estimate their relative pose, especially in large-scale scenarios still remain an open question, because of the characteristic of point cloud mentioned above,.

In this work, we propose an end-to-end method to estimate the rigid transformation between two single-frame point clouds in the absence of initial predictions. The key contributions of our work are as follows: 

First, a self-supervised pre-training method based on metric learning for registration is proposed. 

Secondly, an end-to-end deep-learning approach is proposed by incorporating the self attention and cross attention mechanism to enhance the static structure information and indicate the correspondence of two point clouds. Then virtual corresponding points can be generated and the SVD method can be utilized to solve the registration problem. 

Thirdly, comparison results in ModelNet40 dataset validate that the proposed approach reaches the state-of-the-art. Furthermore, experiments in KITTI dataset demonstrate the potential performance of the proposed approach in large-scale scenarios.

\section{Related Work}

\subsection{Methods to extract point cloud features}
There are three categories of methods to extract features of point cloud.
The first one is to align points to the voxels and then use 3D CNN to extract features \cite{ZhouVoxelNet,SindagiMVX}. The volume representation of point cloud retains the entire information of point cloud. Whereas, due to the sparsity and irregular of the point cloud, the feature extraction process appears to be time-consuming.
The second one is to project 3D point clouds to 2D pixel-based maps and employ 2D CNN \cite{SuMulti}. The typical 2D maps include the spherical map, the camera-plane map and the bird-eye view map. However, in the presence of quantization errors, these approaches are likely to introduce noises and the computational cost remains high. The third way is to directly learn from the raw point cloud. PointNet \cite{Garcia2016PointNet} extracts features of each point independently and uses max-pooling to generate global features. PointNet++ \cite{QiPointNet} further introduces the hierarchical neural network. However, it remains process each point independently, which ignores the local structure information. To solve this, DG-CNN \cite{WuDGCNN} extracts local structure information by respectively using dynamic graph network and the kernel correlation. LPD-net \cite{LiuLPD} mines the local distribution information both in the geometry space and the feature space, which is effective for place recognition in large-scale environments. 

\subsection{Methods for point cloud registration}

The most commonly used approach for point cloud registration is the 
Iterative Closest Point (ICP) \cite{Paul1992A}, the performance of which highly depends on the accuracy of initial estimation information. Unfortunately, the initial value obtained from odometers is not reliable enough and brings about great possibility of local optima. Several methods have proposed to attempt to find a good optima with the ICP method, such as the branch-and-bound (BnB) method \cite{Jiaolong2013Go}, the convex relaxation \cite{MaronPoint}, and the mixed-integer programming \cite{GuMixed}. Nonetheless, these methods have heavy calculation burden and can not meet the requirements of the practical applications. What's more, there is no guarantee of global optima.

There have been several attempts to solve point cloud registration task with learning-based methods. DeepICP \cite{LuDeepICP} uses PointNet++ \cite{QiPointNet} to extract feature and innovatively generate correspondences by learning the matching probabilities among a group of candidates, which can boost the accuracy for sparse point cloud registration. L3-Net \cite{L3net} learns features by PointNet \cite{Garcia2016PointNet}, constructs cost volumes over the solution space and applies CNNS and RNNs to estimate the optimal pose. However, both of the above methods require an initial estimate of the relative pose, which is hard to be obtained accurately in real applications. What's more, as PointNet is used to extract features, the local structure information is neglected. More recently, Deep Closest Point(DCP)\cite{WangDeep} achieves excellent alignment for objects with sharp features and large transformation. Whereas it is infeasible to be applied to large-scale scenes as the point cloud is more sparse, irregular, uneven and the existence of partial correspondences.

\begin{figure}[!b]
	\centering
	\includegraphics[width=0.65\columnwidth]{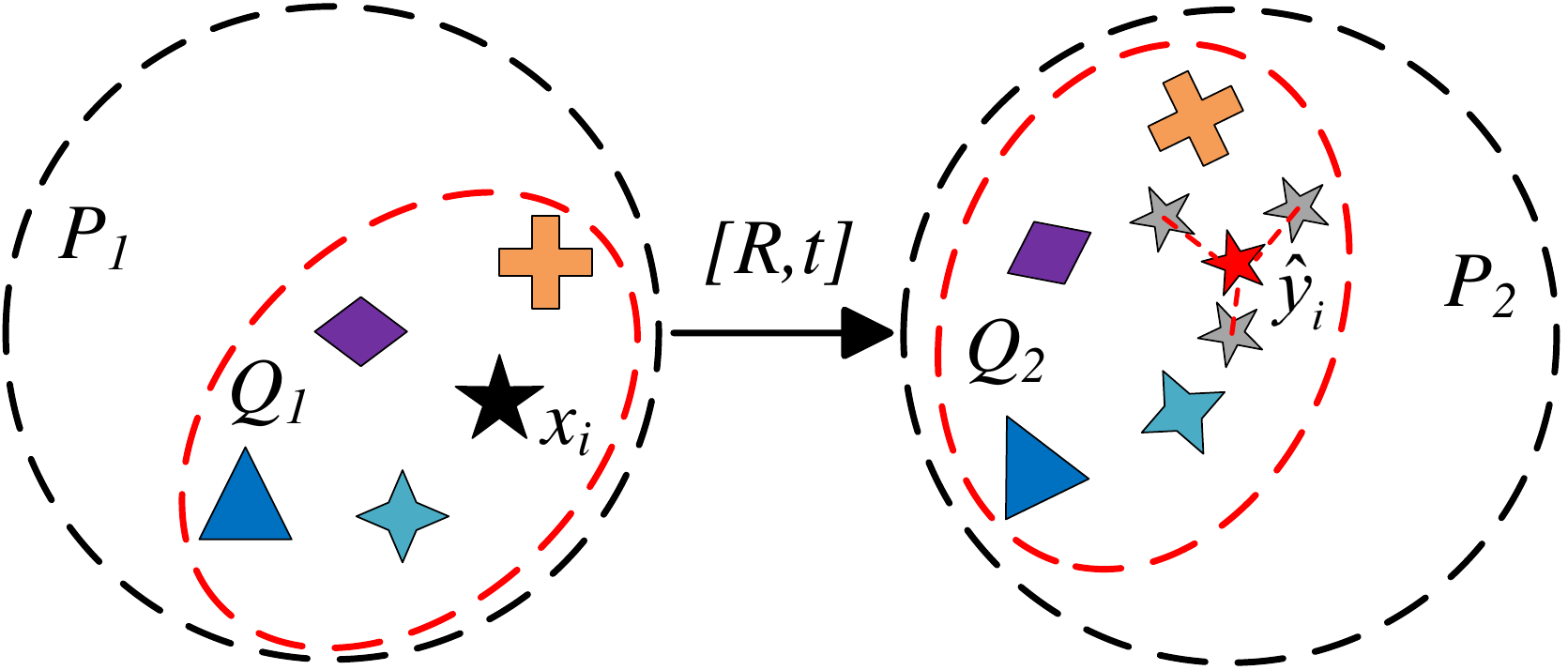}
	\caption{The formulation of the point cloud registration problem.}
	\label{figformulation}
\end{figure}

\begin{figure*}[!t]
	\centering
	\includegraphics[width=2.05\columnwidth]{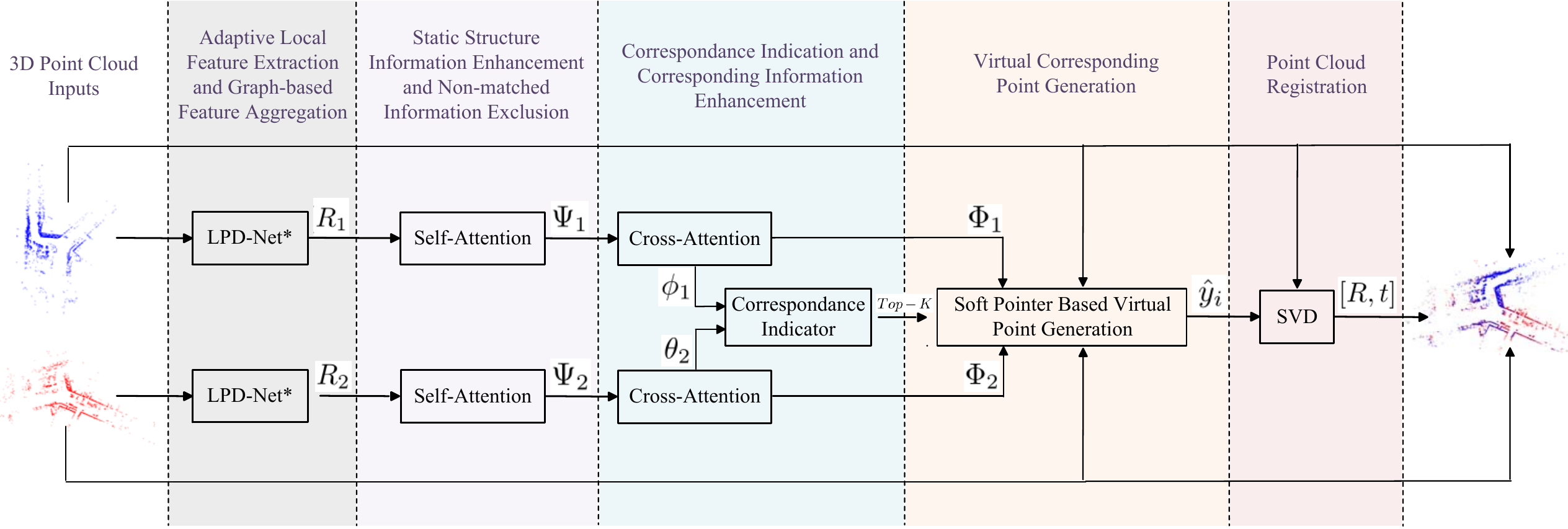}
	\caption{The network architecture of LPD-Registration (LPD-Net* represents the revised version of LPD-Net \cite{LiuLPD}).}
	\label{fignetwork}
\end{figure*}

\begin{figure*}[!t]
	\centering
	\includegraphics[width=2\columnwidth]{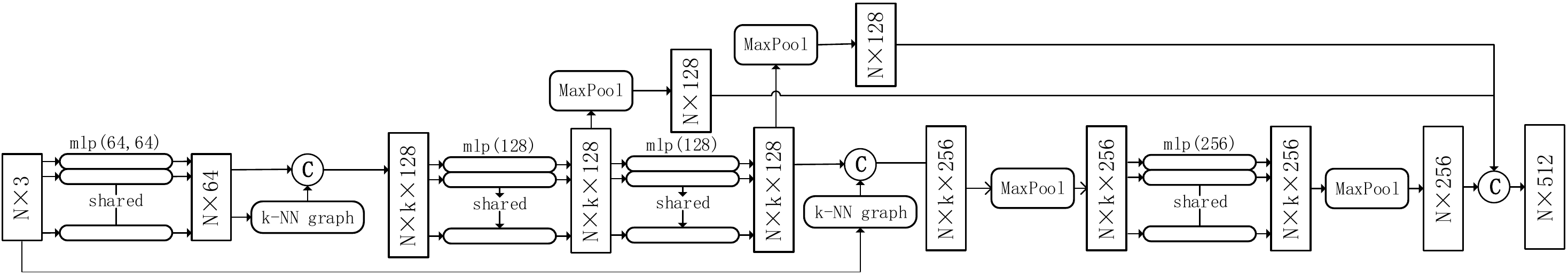}
	\caption{Feature extraction with the revised LPD-Net.}
	\label{figlpd}
	\vspace{-0.1cm}
\end{figure*}
\section{Problem Formulation}

In this section, we formulate the point cloud registration problem. Given two point clouds: $P_1=\{x_i|i=1,...,N\}$ and \[ P_1=\{x_i|i=1,...,N\} \] and $P_2=\{y_i|i=1,...,N\}$, where $x_i$, $y_i\in \mathbb{R}^3$ denote the coordinates of each point. The objective is to calculate the relative transformation between the two point clouds. As shown in Figure \ref{figformulation}, we consider the pose estimation problem of large-scale point clouds with only partially correspondences. The corresponding parts between these two point clouds are represented by $Q_1$ and $Q_2$ respectively.

Due to the sparsity of the point cloud and the perspective variations, the two point clouds do not necessarily have correspondences between their points, which means for a certain point $x_i \in Q_1$, it may be infeasible to find the exact correspondence $y_i\in Q_2$. To solve this problem, an optimal correspondence $\hat{y}_i$ (defined as the virtual corresponding point) will be generated by combining the information of several candidate corresponding points. Then $x_i \in Q_1$ and its virtual corresponding point $\hat{y}_i \in Q_2$ form a correspondence (or a corresponding pair) between the two point clouds.

With the knowledge of a sequence of correspondences between two point clouds, the respective transformation can be obtained by least squares methods. To formulate the problem, the corresponding pairs between two point clouds are denoted by $C_1=\{x_i|i=1,...,K\}$  and $C_2=\{\hat{y}_i|i=1,...,K\}$ (which means $x_i$ and $\hat{y}_i$ are paired). The rigid transformation is denoted as $[R,t]$, where the rotation $R\in SO(3)$ and the translation $T\in \mathbb{R}^3$ . We can obtain the estimation of the transformation by minimizing the mean-squared error:
\begin{equation}
E(R,t)=\frac{1}{K}\sum_1^k||Rx_i+t-\hat{y}_i||^2.
\end{equation}

\section{Network Design}

In this paper, we present a deep-learning based approach, named LPD-Registration (as shown in Figure \ref{fignetwork})), to resolve the pose estimation problem without any initial prediction information. Firstly, a feature extraction network inspired by LPD-Net\cite{LiuLPD} and DenseNet\cite{articledense} is introduced to extract the local features and aggregates them with the k-NN based graph network. Secondly, the self-attention mechanism is used to enhance the static structure information in the large-scale point cloud and exclude the parts which are difficult to match. Thirdly, a cross-attention mechanism is designed to indicate the correspondences between the two input point clouds and thus enhancing the corresponding information. Based on which, the virtual corresponding points can be generated. Here we present two different methods, i.e., the Soft Pointer based and Voting based virtual corresponding point generation. Finally, the rigid transformation $[R,t]$ can be solved by implementing the SVD method to minimize $E(R,t)$.

\subsection{Feature Extraction}

The feature extraction model follows our previous work LPD-Net \cite{LiuLPD}, which solves the place recognition problem by extracting point cloud local structure features and using graph-based feature aggregations. In this paper, some adaptations have been made for the requirement of point cloud registration task. To be specific, we remove the process of aggregating local features to a global descriptor vector, and simplify the network by only performing graph-based neighborhood aggregation in the Cartesian space as well as the feature space, as shown in Figure \ref{figlpd}.

The feature extraction module leverages the direct connection from later layers to the earlier layers to reuse the features and compact the models. To improve the efficiency of our revised LPD-Net, a self-supervised pre-training method is adapted, as shown in
Algorithm\ref{Self-supervised pretraining}. The inputs of the pre-training procedure are the point cloud $P_1$ and a generated point cloud $P_2$ from $P_1$ by a pose transformation. It's worth noting that there is a one-to-one correspondence between each points in $P_1$ and $P_2$ since $P_2$ is generated by using a simple matrix multiplication from $P_1$. The pre-training could help the network learning the spatial structure of the point cloud. Besides, the feature matrix will have the rotation and shift invariance 
benefiting by the generation method of the negative and positive sample pairs(as shown in line\ref{alg1.pos} and line\ref{alg1.neg} in Algorithm\ref{Self-supervised pretraining}).

\begin{algorithm}[h]  
\caption{Self-supervised pre-training algorithm for the revised LPD-Net} 
\label{Self-supervised pretraining} 
\begin{algorithmic}[1] 
\Require 
Point cloud data $P_1$ and $P_2$ 
\Ensure 
The corresponding feature matrix $R_1$ and $R_2$
\State Sample k points from $P_1$ using farthest point sampling(FPS).
\label{alg1:FPS} 
\For{i in k points in $P_1$}
\State Find the corresponding point in $P_2$(with one-to-one correspondence). 
\label{alg1.pos}
\State Append the corresponding point in the positive sample.
\State Find the farthest m points in $P_1$ and get the corresponding point of the farthest point in $P_2$.
\label{alg1.neg}
\State Append the corresponding point in the negative sample.
\EndFor
\State Train the feature extraction network using the triplet loss with $\mathcal{L}_1-norm$ regularization. 
\end{algorithmic} 
\end{algorithm}

\subsection{Self Attention}

Based on the features extracted in the previous module, a self-attention mechanism is introduced for enhancing the static structure information and weakening the parts which are difficult to match in the large-scale point cloud. Specifically, a mask $W_a$ is learned through our self-attention module and a lower weight usually indicates an outer point. Removing the outer points with low weights, points with rich structure information are preserved, which boosts the accuracy of point cloud registration.

The input of the self-attention module is the feature $R_i=\{p_i\}\in \mathbb{R}^{N\times c}$ of all the points in the point cloud $P_1$, where $c$ is the dimensionality of the feature. The self-correlation weight matrix is defined as
\begin{equation}
W_c=Softmax(R_1\cdot R_1^T)\in \mathbb{R}^{N\times N}.
\end{equation}
Then new features $F_1=W_cR_1$ are generated by introducing the self-correlation information. Also, by using mlp regression, the mask $W_a$ can be obtained as:
\begin{equation}
W_a=Softmax(mlp(W_c))\in \mathbb{R}^{N\times 1}.
\end{equation}
Finally, we obtain the new embedding $\Psi_1$ as
\begin{equation}
\Psi_1=\mathrm{f}_{self-attention}(R_1)=W_aF_1.
\end{equation}

The above self-attention mechanism will be also implemented in the point cloud $P_2$ to generate $\Psi_2$.

\subsection{Cross Attention}
Considering the requirements of finding correspondences between two point clouds in the registration task, our method employs cross attention module. Corresponding information is aggregated into the embedding, which promotes the efficiency of finding corresponding parts. Specially, the non-local operation \cite{WangNon} is introduced to realize cross attention. As a result, information of the entire point cloud $P_2$ are included in the feature of each point in point cloud $P_1$ and make it easier to find correspondences.

Take the embedding $\Psi_1$ and $\Psi_2$ of two point clouds as input, we define the following non-local operation:
\begin{equation}
\Pi(\Psi_1,\Psi_2)=Softmax(\theta_2\phi_1^T)V_g\Psi_1,
\end{equation}
where $\theta_2=V_{\theta}\Psi_2$, $\phi_1=V_{\phi}\Psi_1$, $V_{\theta}$, $V_{\phi}$ and $V_g$ are weighted matrices to be learned. The former part of the above equation computes the relationship between $\Psi_1$ and $\Psi_2$, the latter part embeds $\Psi_1$ with a coefficient matrix $V_g$. The idea here is to enhance the corresponding information by investigating the correlation between $\Psi_1$ and $\Psi_2$.

Take the non-local operation as a residual term, new embeddings $\Phi_1$ and $\Phi_2$ are generated by adding the residual term to the previous feature
\begin{equation}
\Phi_1=V_c\Pi(\Psi_1,\Psi_2)+\Psi_1\in \mathbb{R}^{N\times c},
\end{equation}
\begin{equation}
\Phi_2=V_c\Pi(\Psi_2,\Psi_1)+\Psi_2\in \mathbb{R}^{N\times c},
\end{equation}
where $V_c$ is a coefficient matrix.

\subsection{Virtual Corresponding Point Generation}

As illustrated in Section 3, it is infeasible to find the exact match point $y_i\in Q_2$ for each point $x_i\in Q_1$. To tackle the problem, a generator is introduced to produce a virtual corresponding point $\hat{y}_i$ with several candidates.

Since the two point clouds are only partially corresponded, we present a $Top-K$ operation to find $K$ best corresponding points from $Q_1$ and for each one, we find $J_k$ best corresponding candidate points in $Q_2$. More specifically, in $\theta_1\phi_2^T$
\begin{itemize}
\item For each row $g$, find out the maximum value $m_g$ among all columns. The row indices of the $K$ biggest $m_g$ are picked to generate the index $\mathbb{K}$ of the $Top-K$ points in $Q_1$.
\item For each $g\in \mathbb{K}$, find out the $J_k$ biggest elements in the $g_{th}$ row in $\theta_1\phi_2^T$, their column indices are picked to generate the index $\mathbb{J}_g$, which represents the $Top-J_k$ best corresponding points in $Q_2$.
\end{itemize}

With the knowledge of candidate match points $\mathbb{J}_g$, the correspondence $\hat{y}_i$ for each point $x_i$, $i\in \mathbb{K}$ can be estimated by the following method:

\begin{itemize}
\item Soft Pointer based Method: Define $M=\{M_{ij}\}=Softmax(\Phi_1\Phi_2^T)$, for each $i\in \mathbb{K}$,
\begin{equation}
\hat{y}_i=\sum M_{ij}y_j,j\in\mathbb{J}_i.
\end{equation}
\end{itemize}

\subsection{SVD Module}
The final module of our network architecture is to extract the relative pose with the correspondences result obtained in the previous subsection. By using the mean-squared error $E(R,t)$ defined in Section 3, the estimation of $R$ and $t$ can be obtained as
\begin{equation}
[R,t]=\arg\min E(R,t).
\end{equation}

Defined centroids of C1 and C2 as
\begin{equation}
\bar{x}=\frac{1}{K}\sum_{i=1}^{K}x_i,i\in \mathbb{K},
\end{equation}
\begin{equation}
\bar{y}=\frac{1}{K}\sum_{i=1}^{K}\hat{y}_i,i\in \mathbb{K}.
\end{equation}
The cross-covariance matrix $H$ can be obtained as
\begin{equation}
H=\sum_{i=1}^{K}(x_i-\bar{x})(\hat{y}_i-\bar{y}).
\end{equation}

Similar to \cite{LuDeepICP,WangDeep}, the traditional singular value decomposition (SVD) can be used to decompose the matrix $H=USV^T$, then the pose estimation result can be acquired by $R=VU^T$ and $t=-R\bar{x}+\bar{y}$.

\subsection{Loss}

As the direct output of the network, the matching result directly reflects the effectiveness of the network. Therefore, it's essential to introduce the correspondence loss $Loss_{cor}$ by including the accuracy of the virtual correspondences of all the key points:
\begin{equation}
\begin{aligned}
Loss_{cor}&=\frac{1}{N_{cor}}\sum_i||{x_i}-{\tilde{y}_i}||\mathfrak{1}[{ture\_matching}]\\
&+\frac{\gamma}{N_{mis}}\sum_i||{x_i}-{\tilde{y}_i}||\mathfrak{1}[{false\_matching}].
\end{aligned}
\end{equation}
Here, whether a certain correspondence is true or false is decided by the ground-truth rigid transformation. Concretely speaking, the virtual corresponding point $\hat{y}_i$ in point cloud $P_2$ of the key point $x_i$ in $P_1$ is transformed by the ground-truth to obtained $\tilde{y}_i$. 
Aligning the two point clouds with the ground-truth transformation, whether the virtual correspondence is true or false can be determined by testing whether $\tilde{y}_i$ has the nearest spatial distance with $x_i$ in $P_1$ among all the other $N-1$ neighbor points in $P_2$. $N_{cor}$ and $N_{mis}$ denote the number of true matching points and false matching points respectively.\\
It's worth noting that in this work the transformation loss 
$Loss_{trans}$ which is defined in Eq.\ref{trans_loss} is not taken into account in training.
\begin{equation}
\label{trans_loss}
Loss_{trans}=||R^TR^g-I||^2+||t-t^g||^2.
\end{equation}
The transformation loss is a common metric to evaluate the agreement of the ground-truth rigid transformation $[R^g,t^g]$ and the SVD estimation $[R,t]$. However, it would lead to ambiguity in our pose estimation problem. To be specific, the pose estimation process uses the de-centroid coordinates as the input which enlarges the solution space and decreases the convergence efficiency of the algorithm.

Finally, the combined loss is defined as:
\begin{equation}
Loss=\alpha Loss_{cor}+\beta||e||^2.
\end{equation}
Here, the second term is the Tikhonov regularization of the network's parameter $e$, which serves to reduce the complexity of the network. $\alpha$ and $\beta$ are balancing factors.

\section{Experiments}
In this section, we train and evaluate the performance of proposed network on two datasets, ModelNet40 and KITTI, and test it in actual application scenarios. The structure of our network is shown in Figure \ref{fignetwork}. We use revised LPD-Net as the feature extraction layer. 
The value of k in the KNN dynamic graph is 20, and the output dimension of the last layer of the revised LPD-Net is 512. In the Attention layer, the embedding dimension is 1024, and we use layNorm without Dropout. In the virtual corresponding point generation module, we use the voted-based method and select $K=896$ key points in each point cloud and pick up $J_k=32$ candidate matching points in the corresponding point cloud for each key point. All experiments are conducted on two 1080Ti GPU.

\begin{table}[!t]
	\renewcommand{\arraystretch}{1.0}
\caption{Comparison results in ModelNet40.}
	\centering
	\begin{tabular}{c|cc|cc}
		\hline
		\hline
        & \multicolumn{2}{|c|}{Angular Error ($^\circ$)} & \multicolumn{2}{|c}{Translation Error ($m$)}\\
        \hline
		& RMSE & MAE & MSE & MAE\\
		\hline
        ICP(10) & 13.213467 & 7.506543 & 0.064503 & 0.046513\\
		Go-ICP & 11.852313 & 2.588463 & 0.025665 & 0.007092\\
		FGR & 9.362772 & 1.99929 & 0.013939 & 0.002839\\
		PointNetLK & 15.095374 & 4.225304 & 0.022065 & 0.005404\\	
		DCP & 2.545697 & 1.505548 &  0.001763 & 0.001451\\	
		Ours & \bf0.468642 & \bf0.233800 & \bf0.001075 & \bf0.000611\\			
		\hline
		\hline
	\end{tabular}
	\label{tabexp1}
\end{table}

\begin{table}[!t]
	\renewcommand{\arraystretch}{1.0}
\caption{Comparison results in ModelNet40 with additional Gaussian noises.}
	\centering
	\begin{tabular}{c|cc|cc}
		\hline
		\hline
        & \multicolumn{2}{|c|}{Angular Error ($^\circ$)} & \multicolumn{2}{|c}{Translation Error ($m$)}\\
        \hline
		& RMSE & MAE & MSE & MAE\\
		\hline
        ICP(10) & 13.115381 & 7.581575 & 0.065619 & 0.047424\\
		Go-ICP & 11.453493 & 2.534873 & 0.023051 & 0.004192\\
		FGR & 24.651468 & 10.055918 & 0.108977 & 0.027393\\
		PointNetLK & 16.00486 & 4.595617 & 0.021558 & 0.005652\\	
		DCP & 1.08138 & 0.737479 & 0.0015 & 0.001053\\	
		Ours & \bf0.399258 & \bf0.209324 & \bf0.000962 & \bf0.000578\\			
		\hline
		\hline
	\end{tabular}
	\label{tabexp2}
\end{table}

\begin{table}[]
	\renewcommand{\arraystretch}{1.0}
\caption{Comparison results in KITTI.}
	\centering
	\begin{tabular}{c|cc|cc}
		\hline
		\hline
        & \multicolumn{2}{|c|}{Angular Error ($^\circ$)} & \multicolumn{2}{|c}{Translation Error ($m$)}\\
        \hline
		& RMSE & MAE & MSE & MAE\\
		\hline
        ICP(10) & 0.36641 & 0.229963 & 0.491369 & 0.205547\\	
		DCP & 14.035395 & 4.724736 & 0.44365 & 0.301712\\	
		Ours & 0.362187 & 0.218195 & 0.108898 & 0.071794\\
		Ours+ICP &  \bf0.277279 & \bf0.180286 & \bf0.053148 & \bf0.035679\\				
		\hline
		\hline
	\end{tabular}
	\label{tabexp3}
\end{table}

\subsection{Registration Results in the ModelNet40 Dataset}
\subsubsection{Whole-to-whole registration}
In the ModelNet40 dataset, we train our network with 9843 point cloud models and test it with the other 2468 point clouds models. In order to get a point cloud pair, we down-sample each object's point cloud to get a source point cloud with 1024 points. Then we perform rigid pose transformation to obtain the target point cloud: for rotation, the range of the three Euler angles is randomly set from $[0,45^o]$; for translation, the translation is randomly set from $[-0.5m,0.5m]$. 

We compare our proposed approach with several existing approaches, i.e., ICP (with 10 iterations), Go-ICP, FGR, PointNetLK and DCP. For ICP, we refer to the code$\footnote{https://github.com/ClayFlannigan/icp.git}$ on github and extend it to CUDA. For Go-ICP, FGR, PointNetLK, and DCP, we directly use the results provided in \cite{WangDeep}. 
The Mean Squared Error (MSE), the Root Mean Square Error (RMSE) and the Mean Absolute Error (MAE) of rotation drift ($^\circ$) and translational drift ($m$) of all the methods mentioned above are listed in Table \ref{tabexp1}. 
The comparison results show that the proposed approach has reached the state-of-the-art. 

In addition, we also add a Gaussian noise with the distribution of $N(0,0.01)$ into each target point cloud. 
Comparison results in Table \ref{tabexp2} demonstrate the robustness of the proposed approach to external noises. Our proposed approach also reaches the state-of-the-art.

\begin{figure}[!t]
	\centering
	\includegraphics[width=0.8\columnwidth]{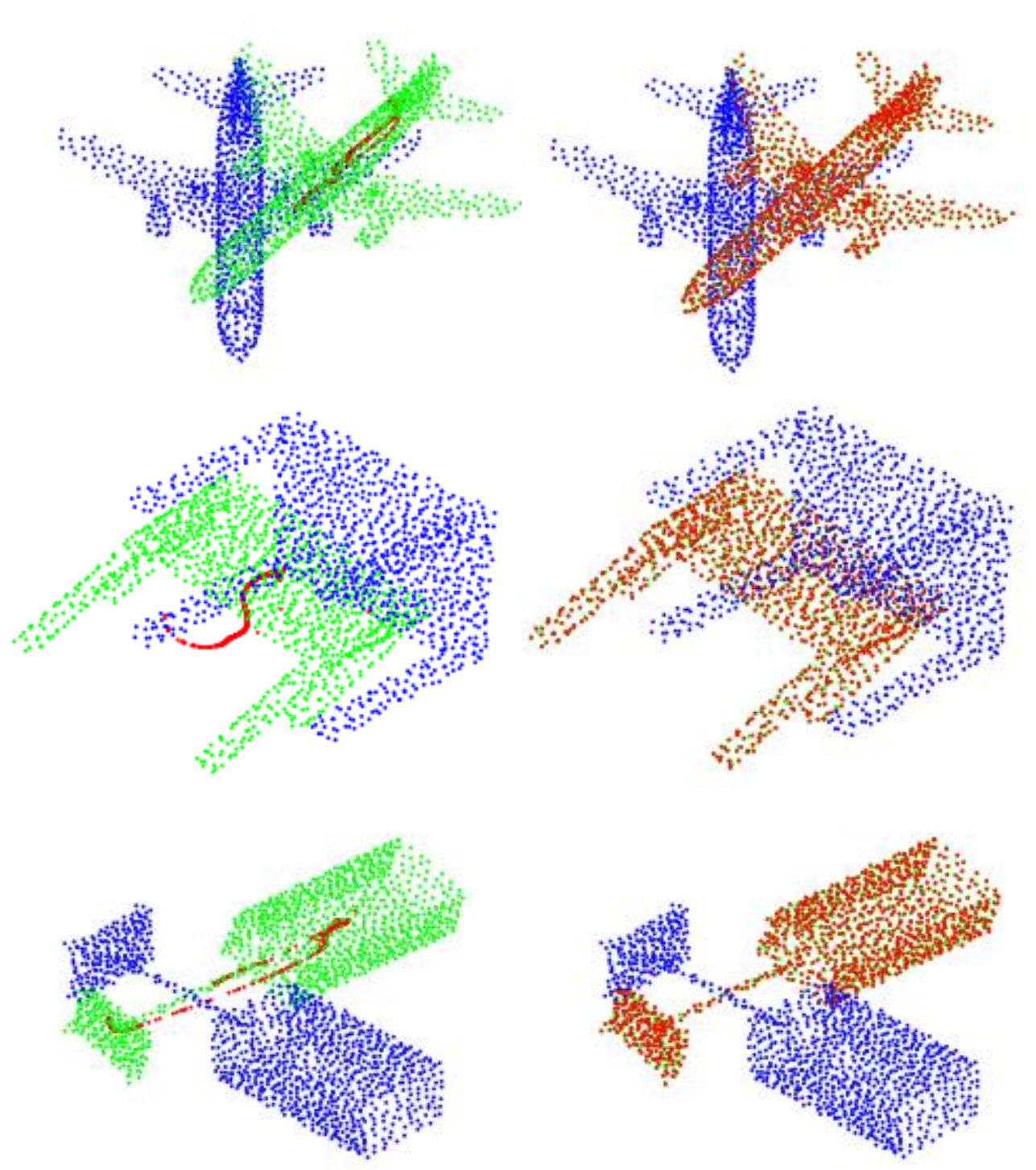}
	\caption{The visualization results of applying different losses. The left is the result of the transformation loss, while the right applies the correspondence loss.}
	\label{cor_visual}
\end{figure}

Finally, in order to prove that removing the transformation loss is helpful for us to generate virtual corresponding points, we show the visualization results of applying only the transformation loss and only the correspondence loss as shown in Figure \ref{cor_visual}. The blue point cloud represents $P_1$, the green point cloud represents $P_2$, and the red point cloud represents the corresponding point cloud of $P_1$ generated by the network. It is worth mentioning that the corresponding points generated by the two kinds of loss training can respectively get a good pose estimation by the SVD, but the corresponding points generated only by the transformation loss are not the corresponding points in the physical sense. It is because this loss cannot apply sufficient constraints on the registration task.

\subsubsection{Partial-to-artial registration}
To evaluate partial-to-partial registration, we refer to the simulation method of \cite{wang2019prnet} by randomly placing a point in space and computing its 768 nearest neighbors in $P_1$ and $P_2$. 

Inspired by PRNet \cite{wang2019prnet}, we also run our network iteratively to get more accurate results. LPD-Pose means to run our network only once, LPD-Pose-ICP means to perform another ICP on the pose transformation results obtained by our network, and LPD-Pose-Iter means to run our network iteratively four times.

The comparison results in Table \ref{tabexpsa} show that our network reach the state-of-the-art in rotation and translation.
\begin{table}[!h]
	\renewcommand{\arraystretch}{1.0}
	\caption{Comparison results in ModelNet40 on unseen point clouds.}
	\centering
	\begin{tabular}{c|cc|cc}
		\hline
		\hline
		& \multicolumn{2}{|c|}{Angular Error ($^\circ$)} & \multicolumn{2}{|c}{Translation Error ($m$)}\\
		\hline
		& RMSE & MAE & RMSE & MAE\\
		\hline
		ICP  & 33.683 & 25.045 & 0.293 & 0.250\\
		Go-ICP \cite{Jiaolong2013Go}  & 13.999 & -0.157 & 0.033 & 0.012\\
		FGR \cite{SindagiMVX}  & 11.238 & 2.832 & 0.030 & 0.008\\
		PointNetLK \cite{yaoki2019pointnetlk}  & 16.735 & -0.654 & 0.045 & 0.025\\	
		DCP \cite{WangDeep}  & 6.709 & 4.448 & 0.027 & 0.020\\	
		PRNet \cite{wang2019prnet}  & 3.199 & 1.454 & \bf0.016 & 0.010\\	
		LPD-Pose & 3.590 & 2.357 & 0.020 & 0.013\\
		LPD-Pose-ICP  & 4.072 & 2.211 & 0.082 & 0.050\\		
		LPD-Pose-Iter & \bf2.097 & \bf0.762 & \bf0.016 & \bf0.006\\			
		\hline
		\hline
	\end{tabular}
	\label{tabexpsa}
\end{table}

\subsection{Registration Results in the KITTI Dataset}

We further train and evaluate the network on the KITTI Dataset to validate the registration performance in large-scale sparse point clouds. The KITTI Dataset contains point clouds captured by a Velodyne HDL-64E Lasers canner together with the ground-truth poses provided by a high-end GPS/IMU Inertial Navigation System. We use Sequence00-02 for training and Sequence03-05 for testing, both for our network and DCP$\footnote{For DCP traning and testing, we use the released code by the authors: https://github.com/WangYueFt/dcp}$.
As the ground is a plane with little cue to pose estimation task, we first remove grounds by RANSAC algorithm. After the ground removal, the point clouds for each frame are down sampled to 1024 points twice in order to obtain point cloud pairs $P_1$ and $P_2$. Then we further perform rigid pose transformation on $P_2$ to obtain the target point cloud: for rotation, the range of the yaw angle is randomly set from $[-30^o,+30^o]$, the range of the other two angles are randomly set from $[-5^o,+5^o]$; for translation, the $x$ and $z$ translations are randomly set from $[-5m,5m]$, the $y$ translations is randomly set from $[-1m,1m]$. 
The comparison results in Table \ref{tabexp3} show that our proposed approach has smaller translation error compared with ICP and DCP. The angular error of the proposed approach is similar with ICP and better than DCP. In additional, with ICP refine, the proposed approach (Ours+ICP) has the best performance. Above results validate the registration performance of the proposed approach in the large-scale point clouds in the presence of local-sparsity and non-correspondence problems.

\section{Conclusion}
In this paper, we present a deep-learning based approach to resolve the pose estimation problem of large-scale point clouds. Specially, the proposed approach does not need any initial prediction information and is robust to local sparsity and partially correspondence problems. Comparison results in ModelNet40 dataset validate that the proposed approach reaches the state-of-the-art and experiments in KITTI dataset demonstrates the potential performance of the proposed approach in large-scale scenarios. 

{\small
\bibliographystyle{ieee_fullname}
\bibliography{egbib}
}

\end{document}